\documentclass[journal]{IEEEtran}

\usepackage[linesnumbered, ruled,vlined]{algorithm2e}
\usepackage{multirow} 
\usepackage{amsmath}
\usepackage{amssymb}
\usepackage{tabularx} 
\usepackage{float}
\usepackage{booktabs} 
\usepackage[table]{xcolor} 
\usepackage{subcaption}
\usepackage{caption}

\usepackage{url} 
\usepackage[colorlinks=true, urlcolor=blue]{hyperref}

\ifCLASSINFOpdf
  \usepackage[pdftex]{graphicx}
  \usepackage{adjustbox}
    \usepackage{tikz}
    \usetikzlibrary{positioning, shapes.geometric, arrows.meta, decorations.pathreplacing}

\else
\fi
\usepackage{array}
\newcolumntype{C}[1]{>{\centering\arraybackslash}m{#1}}
\newcommand{\img}[1]{\includegraphics[width=\linewidth]{#1}}

\begin{document}
%
\title{Noise2Map: End-to-End Diffusion Model for Semantic Segmentation and Change Detection}
%
%
%

\author{Ali~Shibli,~\IEEEmembership{Student Member,~IEEE,}        Andrea~Nascetti,~\IEEEmembership{Member,~IEEE,}
        and~Yifang~Ban,~\IEEEmembership{Senior Member,~IEEE}
\thanks{The authors are within the Division of Geoinformatics, School of Architecture and Built Environment, KTH Royal Institute of Technology, 11428 Stockholm, Sweden (e-mail: shibli@kth.se; nascetti@kth.se; yifang@kth.se).}
}

%
%

\markboth{TGRS-2025-07593}
{Shell \MakeLowercase{\textit{et al.}}: Bare Demo of IEEEtran.cls for Journals}

%



\maketitle

\begin{abstract}

    Semantic segmentation and change detection are two fundamental challenges in remote sensing, requiring models to capture either spatial semantics or temporal differences from satellite imagery. Existing deep learning models often struggle with temporal inconsistencies or in capturing fine-grained spatial structures, require extensive pretraining, and offer limited interpretability—especially in real-world remote sensing scenarios. Recent advances in diffusion models show that Gaussian noise can be systematically leveraged to learn expressive data representations through denoising. Motivated by this, we investigate whether the noise process in diffusion models can be effectively utilized for discriminative tasks. We propose Noise2Map, a unified diffusion-based framework that repurposes the denoising process for fast, end-to-end discriminative learning. Unlike prior work that uses diffusion only for generation or feature extraction, Noise2Map directly predicts semantic or change maps using task-specific noise schedules and timestep conditioning, avoiding the costly sampling procedures of traditional diffusion models. The model is pretrained via self-supervised denoising and fine-tuned with supervision, enabling both interpretability and robustness. Our architecture supports both tasks (SS and CD) through a shared backbone and task-specific noise schedulers. Extensive evaluations on the SpaceNet7, WHU, and xView2 buildings damaged by wildfires datasets demonstrate that Noise2Map ranks on average 1st among seven models on semantic segmentation and 1st on change detection by a cross-dataset rank metric (average F1 primary, IoU tie-break), while being 13$\times$ faster and 3$\times$ smaller than the generative diffusion baseline (DDPM-CD) due to its single-step discriminative inference.
    Ablation studies highlight the robustness of our model against different training noise schedulers and timestep control in the diffusion process, as well as the ability of the model to perform multi-task learning.
\end{abstract}


%
\IEEEpeerreviewmaketitle

\section{Introduction}

\IEEEPARstart{S}{emantic} segmentation (SS) and change detection (CD) are fundamental tasks in remote sensing, enabling critical applications such as environmental monitoring, disaster assessment, and land use analysis. These tasks rely on high-resolution satellite imagery, which presents challenges such as spatial heterogeneity, atmospheric distortion, and temporal inconsistencies. The complexity of bi-temporal data further complicates modeling efforts, making it difficult to extract consistent and reliable spatial or temporal patterns. As a result, CD and SS remain difficult to scale across diverse regions or imaging conditions, particularly when labeled data is limited.

Traditional approaches for SS and CD predominantly rely on convolutional neural networks (CNNs) or Transformer-based architectures. CNNs have been widely adopted for remote sensing tasks \cite{wu2024cmlformer}, but often struggle to capture long-range dependencies and temporal alignment. Transformer-based models \cite{wang2024transformers} address some of these limitations through self-attention but come with high computational costs and require large annotated datasets. 
More recently, diffusion models have emerged as a powerful class of generative models in computer vision, capable of learning complex data distributions through iterative denoising processes \cite{dhariwal2021diffusion, rombach2022high}. Their progressive noise-to-signal refinement enables strong feature learning, even under limited supervision. Remote sensing research has started to explore diffusion models for generative applications, such as cloud removal, super-resolution, and image synthesis \cite{khanna2024diffusionsat, tang2024crs, tian2024swimdiff}. However, most existing methods either focus solely on generation or use diffusion as a feature extractor during the sampling process. Some few recent works attempted discriminative uses of diffusion \cite{Bandara_2025_WACV, luo2024rs}, but they either decouple the denoising process from the downstream task or rely on sampling-based inference. 
Decoupled or sampling-based uses of diffusion learn generic generative features and incur slow, unstable inference, whereas an end-to-end diffusion model back-propagates task losses through the noise schedule so the denoiser learns boundary-aware (SS) and change-sensitive (CD) representations in a single forward pass.

In this paper, we propose \textbf{Noise2Map}, an end-to-end discriminative diffusion model for semantic segmentation and change detection. Unlike prior works, Noise2Map aligns the diffusion denoising trajectory directly with the desired change map or semantic segmentation map.
Through task-specific noise scheduling and end-to-end supervision, our model learns to predict discriminative outputs from noisy inputs without relying on intermediate image reconstruction or handcrafted post-processing. This entirely avoiding the iterative and intensive sampling process required in traditional generative diffusion models.

Beyond performance, Noise2Map also offers an important interpretability benefit. While CNNs and Transformers can be analyzed using post-hoc techniques such as saliency maps or attention visualizations, they lack an inherently transparent prediction process. In contrast, diffusion models allow users to observe how predictions evolve across denoising steps. By visualizing these intermediate representations, our model provides an implicit mechanism to understand how changes or semantic regions emerge, offering a valuable tool for transparency.

\vspace{2mm}
\noindent
In short, our contributions are as follows:
\begin{enumerate}
    \item We propose \textbf{Noise2Map}; an end-to-end discriminative diffusion model that utilizes noise as a discriminator for semantic segmentation and change detection tasks.
    \item We demonstrate that our model achieves strong performance across multiple benchmark datasets (SpaceNet7, WHU, xView2 buildings damaged by wildfire) across the two tasks on seven other state-of-the-art models per task, while requiring significantly less pretraining data.
    \item We conduct ablation studies that show the robustness of our model across different hyperparameter choices and noise schedulers, as well as the ability for the model to multi-task (SS and CD simultaneously).
\end{enumerate}



\section{Related Work}
\label{related_work}


\textbf{Semantic Segmentation in Remote Sensing} 
RS semantic segmentation (SS) underpins land-cover mapping and urban monitoring. Canonical CNN encoder–decoders (U-Net, SegNet) remain strong for dense labeling, while context aggregation with atrous/ pyramid modules (DeepLabv3+, PSPNet) sharpens boundaries and enlarges receptive fields \cite{ronneberger2015u,badrinarayanan2017segnet,chen2017rethinkingatrousconvolutionsemantic,zhao2017pyramid}. Transformer segmentation brings global context with lightweight decoders (e.g., SegFormer) and competitive performance on RS benchmarks; general-purpose decoders (e.g., UPerNet/Mask2Former) are also widely adopted \cite{xie2021segformerr,Xiao_2018_ECCV,cheng2022masked}. RS-specific advances include multi-modal fusion (e.g., RGB+DSM/LiDAR) and weak/self-supervised pretraining to reduce labeling cost (e.g., FTransUNet; SeCo; SSL4EO) \cite{ma2024multilevel,manas2021seasonal,wang2023ssl4eo,cong2022satmae,li2024review,yu2023deep}.

\textbf{Change Detection in Remote Sensing}
Early RS change detection (CD) spans image algebra, post-classification comparison, and object-based analyses \cite{ban2016change}. Modern deep CD largely adopts Siamese CNNs that compare bi-temporal features (e.g., FC-Siamese/UNet variants) but can miss long-range relations and fine boundaries. Transformers improve global context (e.g., BIT \cite{Chen_2022}) and hybrid designs further guide multi-scale fusion using change priors and self-attention (CGNet \cite{han2023change}). Recent transformer or state-space approaches (e.g., ChangeFormer \cite{bandara2022transformer} and ChangeMamba \cite{chen2024changemamba}) push accuracy/efficiency on high-resolution benchmarks. Beyond bi-temporal pairs, \cite{hafner2025continuous} proposes continuous urban CD over time series with temporal feature refinement and multi-task integration, addressing the gap between pairwise CD and city-scale monitoring. These trends motivate architectures that preserve spatial detail, use long-range context, and scale to multi-temporal data.

\textbf{Diffusion Models for Discriminative Tasks}
Diffusion models, initially developed for generative tasks, are increasingly being applied to discriminative tasks. Recent studies have identified specific activations within diffusion models that enhance semantic segmentation and other discriminative tasks, emphasizing the importance of feature selection \cite{meng2025not}. Diffusion-TTA \cite{prabhudesai2023diffusion} refines classifier and segmentor outputs at test-time by leveraging generative feedback, improving robustness in domain shifts. Similarly, diffusion models exhibit zero-shot classification capabilities, enabling classification without explicit training on labeled data, making them useful when annotations are scarce \cite{clark2023text}. They have also been employed for unsupervised semantic correspondence, generating semantic mappings for image matching and transfer learning \cite{hedlin2024unsupervised}, as well as open-vocabulary semantic segmentation, eliminating the need for task-specific fine-tuning \cite{wang2023diffusion}.  
Furthermore, DiffusionDet formulates object detection as a denoising diffusion process from noisy boxes to object boxes \cite{chen2023diffusiondet}. Finally, Diff-Mix enhances image classification by performing inter-class image mixup using diffusion models \cite{wang2024enhance}.
These advancements collectively illustrate how diffusion models are evolving beyond their generative origins, positioning them as powerful tools for discriminative tasks across various domains. 
\\

\textbf{Diffusion in Remote Sensing}
Diffusion models have gained traction in remote sensing for both generative and discriminative tasks, such as image super-resolution, semantic segmentation, and change detection \cite{Yang2022DiffusionMA, 10684806}. While initially developed for generative applications, these models have demonstrated strong feature extraction capabilities that improve performance in classification and segmentation tasks. In change detection, DDPM-CD \cite{Bandara_2025_WACV} employs Denoising Diffusion Probabilistic Models (DDPMs) \cite{ho2020denoising} as feature extractors, showing that diffusion models effectively capture meaningful temporal changes in satellite imagery. DGDM \cite{wan2024leveraging} further enhances urban change detection by integrating a Difference Attention Module and an Image-to-Text adapter, improving accuracy across multiple datasets. Similarly, SiameseMD \cite{Jia2024SiameseMD} combines diffusion models with a Siamese network to better capture bi-temporal differences. In semantic segmentation, 
RS-Dseg \cite{luo2024rs} integrates a diffusion component within a UNet backbone, leveraging unsupervised pretraining and a spatial-channel attention module to enhance efficiency and segmentation performance on high-resolution datasets. Additionally, diffusion models have been utilized for remote sensing super-resolution, with works like \cite{Xiao2023EDiffSRAE, han2023enhancing} demonstrating their ability to enhance spatial resolution. 

This growing adoption of diffusion models in remote sensing highlights its efficiency in learning meaningful representations from satellite imagery. While existing methods typically integrate diffusion models primarily for feature extraction alongside other architectures for SS or CD, we propose \textit{Noise2Map} model that provides a unified, end-to-end diffusion-based framework that directly maps input imagery to output targets.

\section{Method}

\begin{figure*}[ht]
    \centering
    \includegraphics[width=0.76\textwidth]{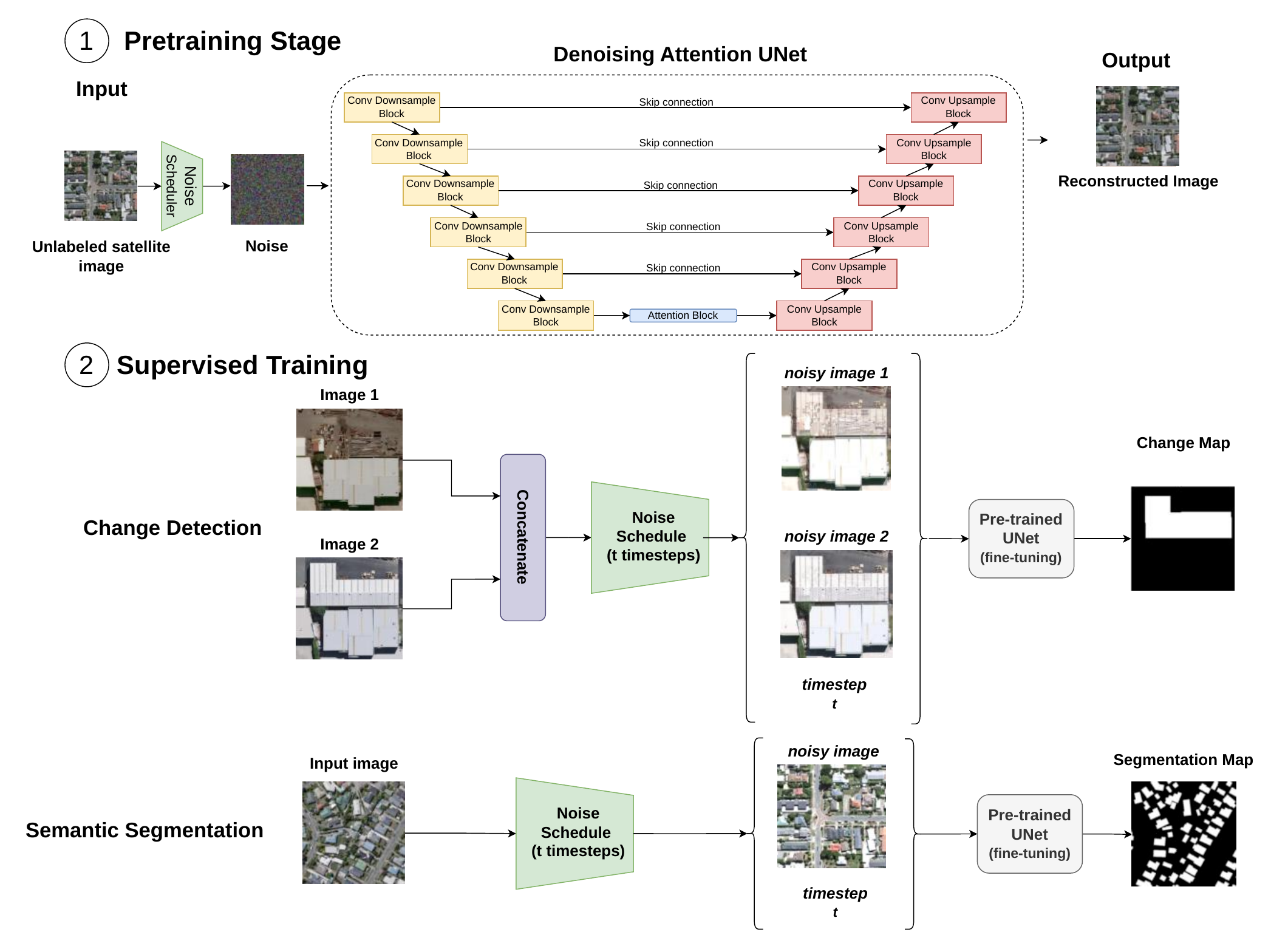}
    
    \caption{Noise2Map overview. 
    (1) Self-supervised pretraining: a denoising attention U-Net is trained on 10k unlabeled satellite images from the AID dataset using standard DDPM objective to reconstruct clean images from noisy inputs. This stage learns general visual representations without task labels.
    (2) Supervised training: the pretrained U-Net is reused and fine-tuned with task-specific structured noise to directly predict outputs for discriminative tasks. For CD, a bi-temporal image pair is used and the noising process gradually transforms the pair to encourage learning change. For SS, a single image is used. In both cases, the model predicts the task output in a single forward pass and is supervised using weighted cross-entropy loss.
    }
    \label{fig:diffusion_model}
\end{figure*}

We propose \textit{Noise2Map}, an end-to-end diffusion model for semantic segmentation and change detection. Unlike prior work that leverages diffusion for generation, we reformulate the diffusion process to learn semantic and change-aware representations directly from the diffusion process via targeted noise scheduling. Noise2Map exploits intermediate noisy representations of input images that encode task-specific features. In other words, we utilize noise from the diffusion process as a discriminative proxy. Figure~\ref{fig:diffusion_model} illustrates our model.

\subsection{Preliminaries and Hypothesis}

The diffusion training process consists of two stages: (a) a forward process in which the input image(s) are progressively noised over $T$ timesteps using a predefined noise scheduler, and (b) a backward process in which a neural network learns to denoise and recover useful structure from the corrupted inputs. 

\textit{\textbf{Hypothesis:} our hypothesis is that noise in the diffusion process can be leveraged not only for image generation, but also as a source of information for discriminative tasks (SS and CD).}


In the following sections, $C$ denotes the channel axis, $H$ the image height, and $W$ the image width, and $\mathcal{D}_{\text{SS}}$ and $\mathcal{D}_{\text{CD}}$ are the noising functions (schedulers) for semantic segmentation and change detection respectively.

\subsection{Diffusion Semantic Segmentation}

Let $\mathbf{x} \in \mathbb{R}^{C \times H \times W}$ denote a satellite image, and let $\mathbf{y} \in \{1, \dots, K\}^{H \times W}$ represent its semantic segmentation mask with $K$ classes. 

We apply a forward diffusion process $\mathcal{D}_{\text{SS}}$ that progressively adds Gaussian noise to the image, while preserving the image at the last noising timestep:

\[
\mathbf{X}^{(T)} = \mathcal{D}_{\text{SS}}(\mathbf{x})
\]

In a way, this creates intermediate noisy representations that the model learns to map directly to the corresponding label (segmentation map). To achieve this, we define a forward noising process over $t = 1, \dots, T$:
\[
\mathbf{x}^{(t)} = \sqrt{\alpha_t} \cdot \mathbf{x} + \sqrt{1 - \alpha_t} \cdot \mathbf{n}^{(t)}, \quad \mathbf{n}^{(t)} \sim \mathcal{N}(0, \mathbf{I})
\]
where $\alpha_t \in [0, 1]$ is a monotonically decreasing variance schedule. At timestep $t = 0$, the image is clean ($\mathbf{x}^{(0)} = \mathbf{x}$), and as $t \rightarrow T$, the intermediate samples become noisier, then noise starts to decrease to get the original image at the last timestep $T$. We design $\alpha_T = 1$ such that at the final timestep, the noised image recovers the original:
\[
\mathbf{x}^{(T)} = \mathbf{x}
\]

During training, we do not rely only on the final timestep. Instead, at each iteration we randomly sample timesteps $t \in \{1, \dots, T\}$ and generate the corresponding intermediate representations $\mathbf{x}^{(t)}$, which are used as inputs to the model alongside the timestep $t$. All noisy inputs share the same ground-truth segmentation mask and are supervised identically. This exposes the model to varying levels of perturbation and encourages robustness and better generalization.

At inference time, however, we use the final timestep $\mathbf{x}^{(T)} = \mathbf{x}$, meaning predictions are made from the clean input image and the timesteps $T$. The diffusion process is therefore used during training to enrich the input distribution, not to corrupt the inference input.

\begin{figure}[h]
  \centering

  \begin{subfigure}{0.8\linewidth}
    \centering

    \begin{subfigure}{0.18\linewidth}
      \centering
      \caption*{\scriptsize $t = 0$}
      \includegraphics[width=\linewidth]{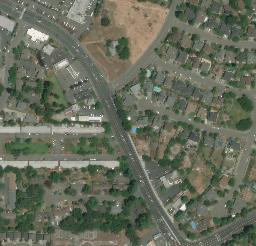}
    \end{subfigure}
    \hfill
    \begin{subfigure}{0.18\linewidth}
      \centering
      \caption*{\scriptsize $t = 250$}
      \includegraphics[width=\linewidth]{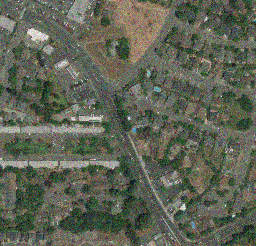}
    \end{subfigure}
    \hfill
    \begin{subfigure}{0.18\linewidth}
      \centering
      \caption*{\scriptsize $t = 500$}
      \includegraphics[width=\linewidth]{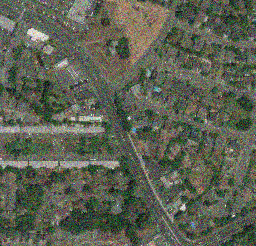}
    \end{subfigure}
    \hfill
    \begin{subfigure}{0.18\linewidth}
      \centering
      \caption*{\scriptsize $t = 750$}
      \includegraphics[width=\linewidth]{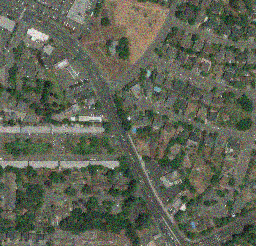}
    \end{subfigure}
    \hfill
    \begin{subfigure}{0.18\linewidth}
      \centering
      \caption*{\scriptsize $t = 999$}
      \includegraphics[width=\linewidth]{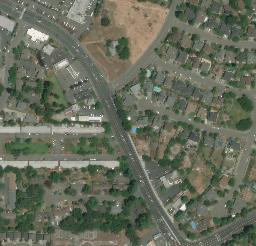}
    \end{subfigure}

  \end{subfigure}

  \vspace{0.3em}

  \begin{subfigure}{0.8\linewidth}
    \centering

    \begin{subfigure}{0.18\linewidth}
      \centering
      \includegraphics[width=\linewidth]{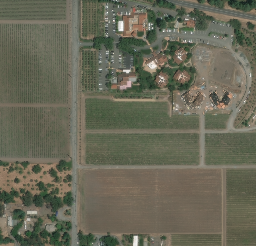}
    \end{subfigure}
    \hfill
    \begin{subfigure}{0.18\linewidth}
      \centering
      \includegraphics[width=\linewidth]{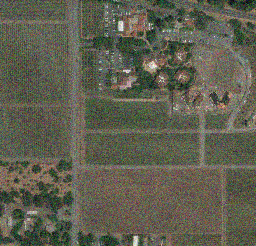}
    \end{subfigure}
    \hfill
    \begin{subfigure}{0.18\linewidth}
      \centering
      \includegraphics[width=\linewidth]{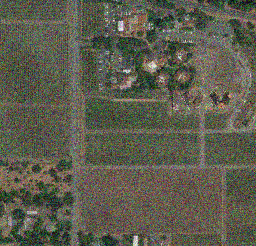}
    \end{subfigure}
    \hfill
    \begin{subfigure}{0.18\linewidth}
      \centering
      \includegraphics[width=\linewidth]{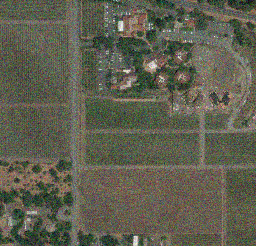}
    \end{subfigure}
    \hfill
    \begin{subfigure}{0.18\linewidth}
      \centering
      \includegraphics[width=\linewidth]{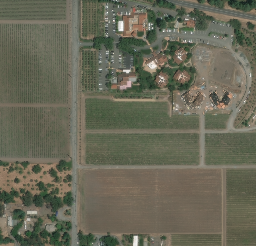}
    \end{subfigure}

  \end{subfigure}

  \caption{Progressive noising-denoising of semantic segmentation input images across timesteps for two examples.}
  \label{fig:ss_noise_progression}
\end{figure}

The noise is introduced smoothly across timesteps allowing semantic structures to gradually emerge as noise is reduced in the reverse process (see Fig.~\ref{fig:ss_noise_progression} for 2 examples on this).

To predict semantic maps, the denoising model $\mathcal{E}_\theta$ takes as input $\mathbf{x}^{(t)}$ and timesteps $t$ and predicts:
\[
\hat{\mathbf{y}}^{(t)} = \mathcal{E}_\theta(\mathbf{x}^{(t)}, t)
\]


This prediction is supervised with the ground true segmentation masks using weighted cross-entropy loss:
\[
\mathcal{L}_{SS} = \text{WCE}(\hat{\mathbf{y}}^{(t)}, \mathbf{y})
\]

By conditioning on intermediate noisy states, the model learns to exploit gradually emerging semantic patterns, refining the mask prediction as noise is removed across denoising steps.

\subsection{Diffusion Change Detection}

Following the same training strategy as in the semantic segmentation setting, the diffusion process is used during training to generate intermediate representations by sampling multiple timesteps, while at inference we use only the final timestep. We focus here on the task-specific formulation for change detection.

Let $\mathbf{x}_{t_1}, \mathbf{x}_{t_2} \in \mathbb{R}^{C \times H \times W}$ denote a bi-temporal image pair at times $t_1$ and $t_2$, respectively. We construct an input tensor $\mathbf{X}^{(0)} = [\mathbf{x}_{t_1}, \mathbf{x}_{t_2}]$ by concatenating both images along the channel axis. $\mathbf{X}^{(0)}$ will be the input to the diffusion change detection model.

During the forward diffusion process, we apply a noise scheduler $\mathcal{D}_{\text{CD}}$ on $\mathbf{X}^{(0)}$. They key in our method is that at the final noising timestep $T$, the resulting tensor $\mathbf{X}$ is transformed into its reversed counterpart:
\[
\mathbf{X}^{(T)} = \mathcal{D}_{\text{CD}}(\mathbf{X}^{(0)}) = [\mathbf{x}_{t_2}, \mathbf{x}_{t_1}]
\]

\begin{figure}[h]
  \centering

  \begin{subfigure}{0.8\linewidth}
    \centering

    \begin{subfigure}{0.18\linewidth}
      \centering
      \caption*{\scriptsize $t = 0$}
      \includegraphics[width=\linewidth]{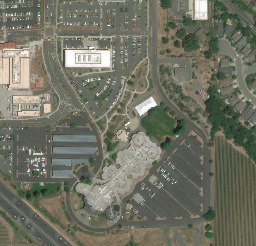}
    \end{subfigure}
    \hfill
    \begin{subfigure}{0.18\linewidth}
      \centering
      \caption*{\scriptsize $t = 250$}
      \includegraphics[width=\linewidth]{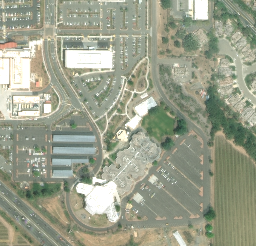}
    \end{subfigure}
    \hfill
    \begin{subfigure}{0.18\linewidth}
      \centering
      \caption*{\scriptsize $t = 500$}
      \includegraphics[width=\linewidth]{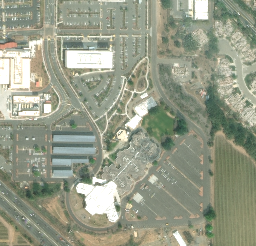}
    \end{subfigure}
    \hfill
    \begin{subfigure}{0.18\linewidth}
      \centering
      \caption*{\scriptsize $t = 750$}
      \includegraphics[width=\linewidth]{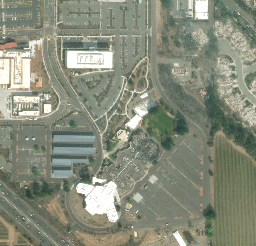}
    \end{subfigure}
    \hfill
    \begin{subfigure}{0.18\linewidth}
      \centering
      \caption*{\scriptsize $t = 999$}
      \includegraphics[width=\linewidth]{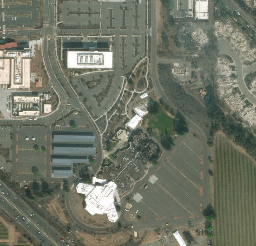}
    \end{subfigure}

    \vspace{0.5em}

    \begin{subfigure}{0.18\linewidth}
      \centering
      \includegraphics[width=\linewidth]{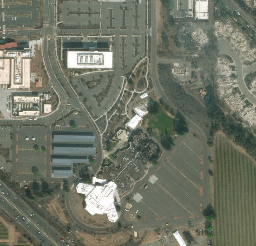}
    \end{subfigure}
    \hfill
    \begin{subfigure}{0.18\linewidth}
      \centering
      \includegraphics[width=\linewidth]{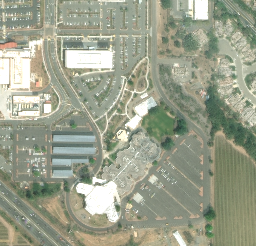}
    \end{subfigure}
    \hfill
    \begin{subfigure}{0.18\linewidth}
      \centering
      \includegraphics[width=\linewidth]{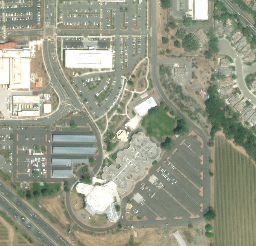}
    \end{subfigure}
    \hfill
    \begin{subfigure}{0.18\linewidth}
      \centering
      \includegraphics[width=\linewidth]{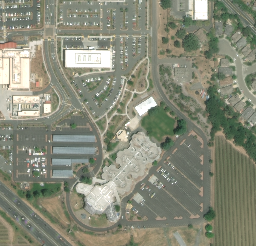}
    \end{subfigure}
    \hfill
    \begin{subfigure}{0.18\linewidth}
      \centering
      \includegraphics[width=\linewidth]{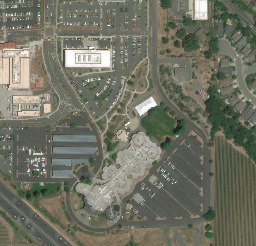}
    \end{subfigure}

  \end{subfigure}


  \begin{subfigure}{0.8\linewidth}
    \centering

    \begin{subfigure}{0.18\linewidth}
      \centering
      \caption*{\scriptsize $t = 0$}
      \includegraphics[width=\linewidth]{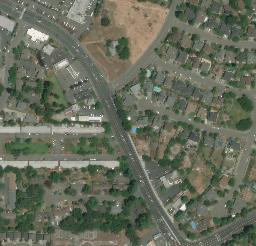}
    \end{subfigure}
    \hfill
    \begin{subfigure}{0.18\linewidth}
      \centering
      \caption*{\scriptsize $t = 250$}
      \includegraphics[width=\linewidth]{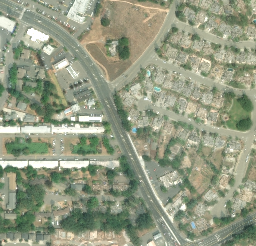}
    \end{subfigure}
    \hfill
    \begin{subfigure}{0.18\linewidth}
      \centering
      \caption*{\scriptsize $t = 500$}
      \includegraphics[width=\linewidth]{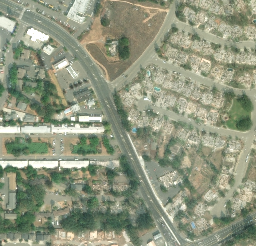}
    \end{subfigure}
    \hfill
    \begin{subfigure}{0.18\linewidth}
      \centering
      \caption*{\scriptsize $t = 750$}
      \includegraphics[width=\linewidth]{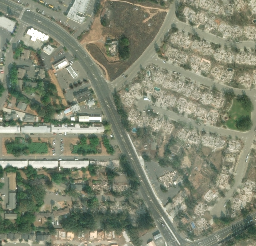}
    \end{subfigure}
    \hfill
    \begin{subfigure}{0.18\linewidth}
      \centering
      \caption*{\scriptsize $t = 999$}
      \includegraphics[width=\linewidth]{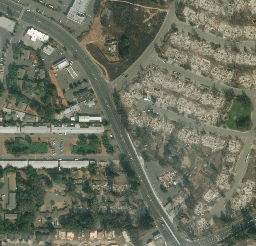}
    \end{subfigure}

    \vspace{0.5em}

    \begin{subfigure}{0.18\linewidth}
      \centering
      \includegraphics[width=\linewidth]{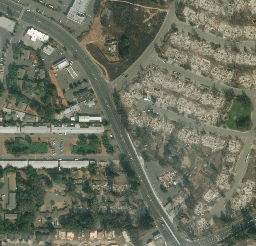}
    \end{subfigure}
    \hfill
    \begin{subfigure}{0.18\linewidth}
      \centering
      \includegraphics[width=\linewidth]{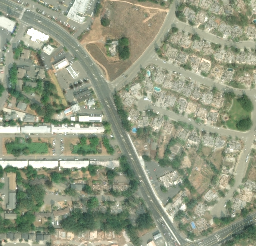}
    \end{subfigure}
    \hfill
    \begin{subfigure}{0.18\linewidth}
      \centering
      \includegraphics[width=\linewidth]{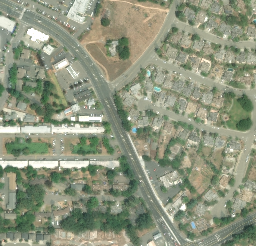}
    \end{subfigure}
    \hfill
    \begin{subfigure}{0.18\linewidth}
      \centering
      \includegraphics[width=\linewidth]{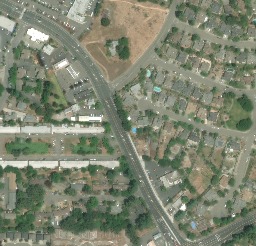}
    \end{subfigure}
    \hfill
    \begin{subfigure}{0.18\linewidth}
      \centering
      \includegraphics[width=\linewidth]{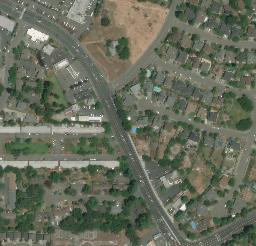}
    \end{subfigure}

  \end{subfigure}

  \caption{Change detection via progressive noising–denoising (two examples). Left column: pre- (top) and post-event (bottom) images; subsequent columns show both views at increasing timesteps, gradually transforming into each other.}
  \label{fig:cd_noise_progression}
\end{figure}

Rather than introducing entirely random noise as in conventional diffusion models, we design the noise to serve as a proxy for learning the transformation between bi-temporal images. To achieve this, we define a forward noising process over $t = 1, \dots, T$:
\[
\mathbf{X}^{(t)} = \sqrt{\alpha_t} \cdot \mathbf{X}^{(0)} + \sqrt{1 - \alpha_t} \cdot \mathbf{N}^{(t)}
\]
where $\alpha_t \in [0, 1]$ is a monotonically decreasing variance schedule and $\mathbf{N}^{(t)}$ is noise designed such that $\mathbf{X}^{(T)} = [\mathbf{x}_{t_2}, \mathbf{x}_{t_1}]$. Importantly, the transition from the original input to the reversed pair is achieved smoothly over timesteps, ensuring a gradual and continuous transformation rather than an abrupt swap at $t = T$ (see Fig.~\ref{fig:cd_noise_progression} for 2 examples on this).


\textbf{Remark}: (1) \textit{Physically}, this noise scheduling simulates a continuous temporal evolution or ``morphing'' between the pre and post states. By forcing the model to observe and process these smooth, intermediate representations, it learns the underlying dynamics of how the landscape transitions over time, rather than simply memorizing hard static pixel differences.
(2) \textit{Mathematically}, this formulation offers an advantage over conventional change detection models that rely on symmetric difference operations (e.g., $|x_{t_1} - x_{t_2}|$), which inherently discard the directionality of the change. By driving the variance schedule to directionally interpolate from $[\mathbf{x}_{t_1}, \mathbf{x}_{t_2}]$ to $[\mathbf{x}_{t_2}, \mathbf{x}_{t_1}]$, the intermediate representations break temporal symmetry. Consequently, the noise signal at any given timestep $t$ encodes the temporal trajectory (the vector of change) rather than just the magnitude. This asymmetric conditioning equips the denoising network to effectively capture and distinguish direction-dependent variations (e.g., a building being constructed versus a building being demolished).

To predict the change maps, the denoising model ($\mathcal{E}_\theta$) takes the images $\mathbf{X}^{(t)}$ and the timesteps used during the transformation ($t$) and predicts the additive noise, which is the change map in our case:
\[
\hat{\mathbf{C}}^{(t)} =  \mathcal{E}_\theta(\mathbf{X}^{(t)}, t)
\]

The prediction is supervised using a weighted cross-entropy loss against the true binary change mask $\mathbf{C} \in \{0,1\}^{1 \times H \times W}$:
\[
\mathcal{L}_{CD} = \text{WCE}(\hat{\mathbf{C}}^{(t)}, \mathbf{C})
\]

In other words, the denoising model is predicting the change map using the noise signal from the noising timesteps in the forward process. This formulation allows the model to treat temporal change as a structured noise signal within the denoising trajectory. 

\subsection{Denoising Model}

The denoising model we employ is an attention denoising UNet model \cite{ho2020denoising} with five downsampling and upsampling blocks. The encoder consists of convolutional layers with channel dimensions (3–128–256–512), while the decoder mirrors this structure. A ResNet block and a self-attention module at the bottleneck enhance global context modeling. Timestep embeddings are encoded using a sine function and passed through two fully connected layers with 512 hidden units. It is illustrated in part 1 of Figure~\ref{fig:diffusion_model}.

We note that the noise schedulers are fixed and non-learnable; all gradient updates are applied to the denoising UNet during both pretraining and task-specific training.

\section{Experiments}
\label{experiments}

In this section, we describe our experimental setup and results. The main results reported are on three public datasets: Spacenet7, WHU, and xView2 buildings damaged by wildfire, both their semantic segmentation and change detection counterparts. The analysis is organized into the following subsections: data description, experimental setup, comparison of methods, and ablation studies.

\subsection{Datasets}


We conduct experiments on three benchmark datasets for SS and CD. Each dataset presents unique challenges, ranging from urban building and structural change detection to wildfire-affected area segmentation. The datasets were selected based on their diverse applications, coverage of both temporal and spatial changes, and varied sizes to validate model robustness across different data sizes. To ensure consistency across experiments, all images are cropped to $256\times256$ pixels, RGB format, and normalized according to dataset-specific configurations. The datasets are:

\begin{enumerate}
    \item SpaceNet7 Dataset \cite{van2021multi} is primarily structured for building footprint segmentation, but we also adopted the train/validation/test splits from \cite{hafner2023semi} for change detection. The dataset includes 9752 bi-temporal image pairs for change detection and 19504 semantic maps for segmentation tasks.

    \item Buildings damaged from wildfire subset of XView2 Dataset \cite{lam2018xview}, the smallest of the three, which includes 150 image pairs for detecting building damage from wildfires (CD) and 291 masks for segmenting wildfire-affected regions (SS).
    
    \item WHU Building Dataset \cite{ji2018fully}, which initially contained several redundant images with no corresponding masks, was refined by removing some of these instances, resulting in a final set of 959 image pairs for urban building change detection (CD) and 8188 masks for building segmentation in urban areas (SS).
    
\end{enumerate}

\subsection{Experimental Setup and Implementation Details}
\label{training_setup}


\textbf{Pretraining Phase.} We use a self-supervised denoising objective where the denoising UNet learns to reconstruct images from Gaussian noise using DDPM scheduler over $T = 1000$ timesteps. The model is pre-trained using AdamW optimizer (learning rate $1 \times 10^{-4}$ with cosine decay and warmup) and Mean Squared Error (MSE) loss. Pretraining is conducted for 200 epochs using 8 NVIDIA RTX 3080Ti GPUs.

The purpose of this pretraining stage is not to solve the downstream tasks directly, but to learn general-purpose visual representations from large amounts of unlabeled satellite imagery. 
This provides a strong initialization for the subsequent supervised stage, improving convergence, stability, and generalization. We further include an ablation study on the effect of the pretraining dataset in section \ref{effect_of_pretraining_subsection}.


\textbf{Pretraining Dataset.} We pre-train the denoising UNet model on a subset of $10{,}000$ images sampled from the AID dataset~\cite{xia2017aid}, which provides diverse high-resolution aerial imagery across multiple land-use and scene categories. We select three spectral bands—[B2, B3, B4], corresponding to the Blue, Green, and Red channels—to ensure compatibility with our RGB-based downstream datasets. Each input image is cropped to \(256 \times 256\) pixels.

\textbf{Training Phase.} 
We train the pretrained model for each task using a weighted Cross Entropy loss to address class imbalance. For change detection, we apply a 3:1 (change:no-change) weighting ratio on xView2-Wildfire and SpaceNet7-CD, and a 1:1 ratio on WHU-CD, which exhibits a more balanced label distribution. For semantic segmentation, we use 3:1 on xView2-Wildfire and WHU, and 5:1 on SpaceNet7, guided by each dataset's class frequencies. In practice, these weights were chosen directly from class imbalance statistics and did not require extensive manual tuning.
Although we experimented with alternative loss functions, including weighted RMSE and focal loss, weighted Cross Entropy provided the best results across both tasks.
Furthermore, we conducted experiments to assess whether boundary-aware supervision improves Noise2Map. We implemented standard pixel-wise boundary reweighting losses commonly used in semantic segmentation and change detection, including (i) fixed boundary reweighting using morphological boundary maps and (ii) radius-based boundary emphasis with different boundary weights. In our experiments, these losses did not improve performance and in several cases degraded both F1 and IoU. 
These results suggest that explicitly enforcing local boundary constraints may conflict with the diffusion-conditioned discriminative objective, which learns structural consistency implicitly through denoising.

\textbf{Code Implementation.} All models are implemented in PyTorch in Python. Models are trained with mixed-precision and gradient scaling. We use the Adam optimizer with a learning rate of $1 \times 10^{-4}$ and a gradient accumulation factor of 2. For our experiments, we use a batch size between 10 and 20 across multiple NVIDIA RTX 3080 GPUs and train up to 200 epochs. Unless otherwise noted, the main noise scheduler in the experiments is DDIM with $T = 1000$ steps. The implementation and training scripts will be made publicly available at \textbf{\url{https://github.com/alishibli97/noise2map}}.

\textbf{Eval Metrics.} We report the evaluation metrics including precision, recall, F1-score, and IoU,
in all experiments. Ablation studies further explore the effect of noise scheduler choice, diffusion steps, and task coupling.


\subsection{Comparison Methods}\label{sec:comparison_methods}

To thoroughly evaluate Noise2Map, we benchmark against seven state‑of‑the‑art semantic segmentation models and seven change detection models spanning diverse architectures including convolutional, transformer-based, diffusion-based, and state‑space/Mamba networks. This selection ensures fair, representative comparison and highlights the complementary strengths across model families.

Semantic Segmentation:
\begin{itemize}
    \item \textbf{UNet} \cite{ronneberger2015u} – A classic encoder–decoder convolutional architecture widely used for SS, known for its skip connections that help preserve spatial information.
    
    \item \textbf{UNet++} \cite{zhou2018unet++} – An enhanced version of UNet with nested and dense skip connections, improving segmentation accuracy by refining feature fusion.
    
    \item \textbf{DeepLabV3+} \cite{chen2017rethinkingatrousconvolutionsemantic} – Incorporates spatial pyramid pooling and an encoder–decoder structure to capture rich multi-scale contextual information.
    
    \item \textbf{SegFormer} \cite{xie2021segformerr} – A transformer-based model that combines lightweight MLP decoders with hierarchical vision transformer encoders, offering a strong balance between accuracy and efficiency.

    \item \textbf{UPerNet} \cite{Xiao_2018_ECCV} – A unified framework built on FPN and PSPNet principles, leveraging multi-scale features for accurate scene parsing.
    
    \item \textbf{DPT} \cite{ranftl2021vision} – A dense prediction transformer combining a Vision Transformer backbone pretrained with DINO (self-supervised learning) and a convolutional decoder, enabling high-resolution semantic segmentation with strong generalization capabilities.
    
    \item \textbf{RS3Mamba} \cite{ma2024rs3mambavisualstatespace} – A recent segmentation model based on Mamba, a structured state-space architecture capable of long-range spatial reasoning with improved efficiency.
    
    \end{itemize}

Change Detection:
\begin{itemize}
    \item \textbf{FC-SiamConc} \cite{daudt2018fully} – A dual-branch U-Net architecture (Siamese) that processes bi-temporal inputs independently before fusing their features, known for its simplicity and effectiveness in early CD tasks.
    
    \item \textbf{BIT} \cite{Chen_2022} – Bitemporal Image Transformer that tokenizes paired images and applies cross-temporal attention to capture long-range spatial-temporal dependencies for CD.
    
    \item \textbf{ChangeFormer} \cite{bandara2022transformer} – A Transformer-based framework with a hierarchical encoder and MLP decoder, designed to capture multi-scale context and refine predictions across resolution levels.
    
    \item \textbf{CGNet-CD} \cite{han2023change} – A CNN-based model integrating self-attention mechanisms to enhance the clarity of change boundaries and suppress noise within unchanged areas.
    
    \item \textbf{ELGC-Net} \cite{noman2024elgcnetefficientlocalglobalcontext} – A context-enhanced model that combines local and global cues with edge-aware guidance to sharpen prediction at object boundaries.
    
    \item \textbf{DDPM-CD} \cite{Bandara_2025_WACV} – Denoising diffusion probabilistic model for CD, establishing a generative baseline that highlights the benefits of progressive refinement in predictions.
    
    \item \textbf{MambaBCD} \cite{chen2024changemamba} – A Mamba state-space model, designed to capture sequential and spatial patterns in bi-temporal satellite imagery.
\end{itemize}

All models are implemented via either official code or the Segmentation Models PyTorch (SMP) library, using default settings from their original papers. This ensures a fair comparison across various model families and architectures. 
For fair comparison, all baseline models were initialized using the standard pretrained weights provided in their official implementations or widely used libraries, following the configurations reported in their original papers. Many baselines rely on ImageNet-pretrained backbones or publicly released checkpoints that were further trained on large-scale remote sensing datasets. For example, the DDPM-CD model is initialized from a checkpoint pretrained on the Million-AID dataset. In contrast, Noise2Map is pretrained on a subset of only 10k images from the AID aerial scene dataset. Therefore, several baseline models benefit from substantially larger-scale pretraining (e.g., ImageNet with millions of images or Million-AID), whereas our model uses a much smaller pretraining dataset. Importantly, we follow the standard pretrained weights used in prior literature to ensure reproducibility and consistency with previously reported results rather than retraining baselines under modified pretraining conditions.

\subsubsection{Quantitative Analysis}

We evaluate Noise2Map on three datasets and two tasks against seven strong baselines per task. As shown in Table~\ref{tab:cdss_all_metrics}, Noise2Map delivers state-of-the-art or highly competitive performance across all settings.

\textbf{Rank aggregation.} We add a summary Rank column that aggregates performance across datasets per task by averaging each model’s per-dataset ranks on \emph{F1} (higher is better) with mean \emph{IoU} as a tie-breaker; the average rank is converted to an ordinal score (1 = best; lower is better). Under this criterion, \textbf{Noise2Map} ranks \textbf{1st} on both tasks.



\begin{table*}[t]
\centering
\caption{Comparison of Noise2Map and SOTA models on SpaceNet7, WHU, and XView2 buildings damaged by wildfire datasets using precision, recall, F1 score, and IoU. Best results are in \textbf{bold} and second best are \underline{underlined}. The final \textbf{Rank} aggregates performance across datasets (lower is better).}
\small
\renewcommand{\arraystretch}{1.15}
\setlength{\tabcolsep}{4pt}
\resizebox{0.75\textwidth}{!}{
\begin{tabular}{l|c|cccc|cccc|cccc|c}
\toprule
\rowcolor[HTML]{E0E0E0}
\multicolumn{15}{l}{\textbf{\hspace{0.3cm} Semantic Segmentation}} \\
\midrule
\multirow{2}{*}{\textbf{Model}} &
\multirow{2}{*}{\shortstack{\textbf{Params}\\\textbf{(M)}}} &
\multicolumn{4}{c|}{\textbf{SpaceNet7-SS}} &
\multicolumn{4}{c|}{\textbf{WHU-SS}} &
\multicolumn{4}{c|}{\textbf{XView2-wildfire-buildings-SS}} &
\multicolumn{1}{c}{\textbf{Rank}} \\   
& & Prec. & Rec. & F1 & IoU & Prec. & Rec. & F1 & IoU & Prec. & Rec. & F1 & IoU & \\ 
\midrule

UNet           &
32.5 &
68.94 & 78.67 & \underline{71.53} & \underline{61.94} &
88.10 & 96.08 & 91.51 & 84.91 &
78.76 & \underline{83.56} & 80.92 & 71.01 & 3
\\

UNet++         &
49.0 &
69.62 & 77.81 & 71.52 & 61.88 &
90.69 & \underline{96.93} & 93.47 & 88.11 &
82.52 & 81.94 & \underline{82.23} & \underline{72.65} & 2
\\

DeepLabV3+     &
26.7 &
68.93 & 74.20 & 69.94 & 60.53 &
87.72 & 96.39 & 91.36 & 84.67 &
77.84 & 78.68 & 78.25 & 68.11 & 7
\\

SegFormer      &
5.6 &
66.18 & 77.42 & 68.63 & 59.08 &
89.79 & 96.87 & 92.89 & 87.14 &
78.33 & 83.52 & 80.64 & 70.67 & 4
\\

UPerNet        &
37.3 &
64.66 & \textbf{82.50} & 67.73 & 57.42 &
89.23 & 96.48 & 92.38 & 86.32 &
80.65 & 80.60 & 80.63 & 70.77 & 6
\\

DPT            &
121.0 &
58.19 & 73.45 & 58.12 & 48.76 &
86.61 & 95.24 & 90.21 & 82.89 &
78.75 & 75.63 & 77.07 & 66.96 & 8
\\

RS3Mamba  &
43.3 &
\textbf{82.79} & 58.57 & 62.60 & 55.25 &
\textbf{96.18} & 92.73 & \underline{94.36} & \underline{89.63} &
\textbf{89.14} & 77.66 & 80.34 & 70.70 & 5
\\

\midrule
Noise2Map-SS  &
113.7 &
\underline{70.33} & \underline{78.83} & \textbf{72.02} & \textbf{62.45} &
\underline{94.86} & \textbf{97.75} & \textbf{95.69} & \textbf{92.90} & 

\underline{86.44} & \textbf{87.38} & \textbf{86.90} & \textbf{78.55} & \textbf{1} \\

\arrayrulecolor{gray}\midrule[0.5pt]\arrayrulecolor{black}

\rowcolor[HTML]{E0E0E0}
\multicolumn{15}{l}{\textbf{\hspace{0.3cm} Change Detection}} \\
\midrule
\multirow{2}{*}{\textbf{Model}} &
\multirow{2}{*}{\shortstack{\textbf{Params}\\\textbf{(M)}}} &
\multicolumn{4}{c|}{\textbf{SpaceNet7-CD}} &
\multicolumn{4}{c|}{\textbf{WHU-CD}} &
\multicolumn{4}{c|}{\textbf{XView2-wildfire-buildings-CD}} &
\multicolumn{1}{c}{\textbf{Rank}} \\   
& & Prec. & Rec. & F1 & IoU & Prec. & Rec. & F1 & IoU & Prec. & Rec. & F1 & IoU & \\
\midrule

FCSiamConc     &
43.7 &
50.88 & 50.39 & 48.02 & 45.50 &
93.95 & 94.31 & 94.13 & 89.14 &
85.35 & \underline{80.87} & 82.92 & 73.55 & 5
\\

BIT            &
11.9 &
49.83 & 50.38 & 48.01 & 45.47 &
95.23 & 94.06 & 94.63 & 90.01 &
81.76 & 69.00 & 73.44 & 63.61 & 7
\\

ChangeFormer-v6   &
41.0 &
54.06 & 50.36 & 48.04 & 45.53 &
94.01 & 92.50 & 93.23 & 87.63 &
88.12 & 75.17 & 80.06 & 70.36 & 6
\\

CGNet-CD       &
39.0 &
49.08 & 50.30 & 47.88 & 45.40 &
\textbf{97.11} & 95.23 & \textbf{96.14} & \textbf{92.67} &
\textbf{93.26} & 79.32 & \underline{84.71} & \underline{75.84} & 3
\\

ELGC-Net       &
10.6 &
49.08 & 50.30 & 47.88 & 45.40 &
88.50 & 92.20 & 90.18 & 82.65 &
75.36 & 80.35 & 77.56 & 67.22 & 8
\\

DDPM-CD       &
437.5 &
\underline{73.56} & \underline{66.14} & 68.91 & 58.91 &
\underline{96.59} & \textbf{95.68} & \underline{96.13} & \underline{92.65} &
\underline{91.96} & 78.59 & 83.77 & 74.67 & \textbf{2}
\\

MambaBCD-Base  &
92.4 &
\textbf{78.65} & 65.14 & \underline{69.46} & \underline{60.46} &
92.30 & \underline{95.46} & 93.77 & 88.52 &
86.52 & 80.21 & 83.00 & 73.66 & 4
\\

\midrule
Noise2Map-CD &
113.7 &
70.24 & \textbf{77.00} & \textbf{71.43} & \textbf{61.91} &

95.94 & 94.64 & 95.27 & 90.39 &
88.43 & \textbf{85.52} & \textbf{86.91} & \textbf{78.59} & \textbf{1}
\\
\bottomrule

\end{tabular}
}
\label{tab:cdss_all_metrics}
\end{table*}

For \textbf{semantic segmentation}, Noise2Map achieves the best overall F1 and IoU on WHU (F1 = 95.69, IoU = 92.90) and XView2-wildfire (F1 = 86.90, IoU = 78.55), improving over the widely used UNet baseline by +4.18 F1 / +7.99 IoU on WHU and +5.98 F1 / +7.54 IoU on XView2. On SpaceNet7, the margin versus UNet is narrower (62.45 vs.\ 61.94 IoU), likely due to simpler structures where UNet already captures coarse boundaries effectively. Still, Noise2Map matches or exceeds competitive baselines—including ImageNet-1K–pretrained models—while using only modest pretraining on 10k AID patches. RS3Mamba attains very strong precision on WHU and XView2 but with lower recall, whereas Noise2Map balances precision and recall across datasets. Overall, this places Noise2Map as the top-ranked model (Rank 1) for SS.

For \textbf{change detection}, Noise2Map achieves the best overall performance across datasets. On SpaceNet7-CD, it attains the highest F1 (71.43) and IoU (61.91), and also achieves the highest recall (77.00), outperforming both discriminative and diffusion baselines.
On XView2 wildfire building damage detection, Noise2Map similarly achieves the best F1 (86.91) and IoU (78.59), while obtaining the highest recall (85.52), indicating strong sensitivity to subtle structural changes.
On WHU-CD, Noise2Map remains highly competitive (F1 = 95.27, IoU = 90.39), though slightly below CGNet-CD and DDPM-CD. We hypothesize that this difference is partly due to architectural specialization: CGNet-CD is designed specifically for change detection with explicit change-prior modeling and attention mechanisms, whereas our diffusion denoiser is shared across both CD and SS tasks to support both tasks. Despite this, Noise2Map achieves the best overall ranking across datasets, demonstrating strong cross-dataset robustness and generalization.
Furthermore, Noise2Map training and inference strategy is much more efficient than the DDPM\textendash CD (the other diffusion change detection model that ranks second overall): 
: 3$\times$ less FLOPs, 3.85$\times$ less params, 13.49$\times$ faster.
We note that this efficiency advantage largely arises from the different inference paradigms of the two approaches. DDPM-CD follows a traditional generative diffusion framework, which requires iterative denoising steps during sampling, whereas Noise2Map reformulates diffusion into a single-step discriminative prediction task. Therefore, the comparison mainly highlights the practical efficiency benefits of the proposed formulation rather than a direct algorithmic equivalence between the two approaches.

\textbf{Overall,} Noise2Map delivers consistent improvements over strong baselines, ranks \textbf{1st} on both SS and CD.
These results confirm that diffusion noise serves as an effective discriminative supervisory signal without sacrificing efficiency.

\subsubsection{Qualitative Analysis}

\begin{figure*}[h]
    \centering
    \setlength{\tabcolsep}{1pt}
    \renewcommand{\arraystretch}{0.9}

    \begin{subfigure}{0.7\textwidth}
        \centering
        \resizebox{\textwidth}{!}{
            \begin{tabular}{cccccccccccc}
                \Huge Image & \Huge UNet & \Huge UNet++ & \Huge DeepLabV3+ & \Huge SegFormer & \Huge UPerNet & \Huge DPT & \Huge RS3Mamba & \Huge Noise2Map (ours) & \Huge Ground Truth \\

                \includegraphics[width=0.5\textwidth]{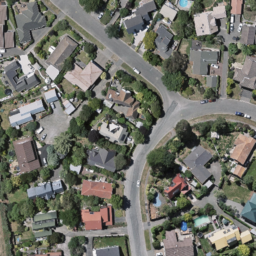} &
                \includegraphics[width=0.5\textwidth]{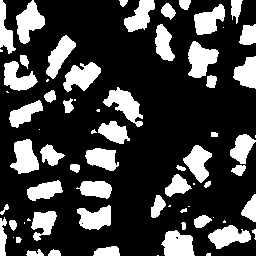} &
                \includegraphics[width=0.5\textwidth]{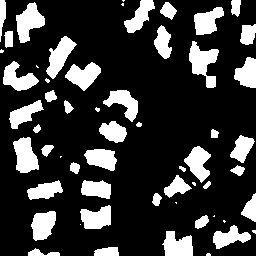} &
                \includegraphics[width=0.5\textwidth]{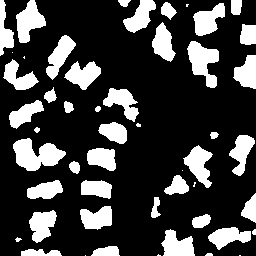} &
                \includegraphics[width=0.5\textwidth]{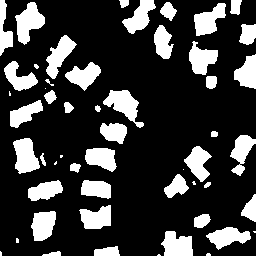} &
                \includegraphics[width=0.5\textwidth]{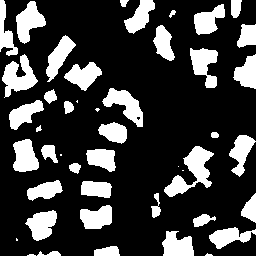} &
                \includegraphics[width=0.5\textwidth]{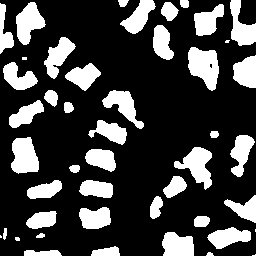} &
                \includegraphics[width=0.5\textwidth]{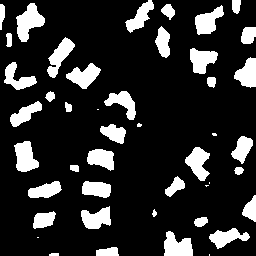} &
                \includegraphics[width=0.5\textwidth]{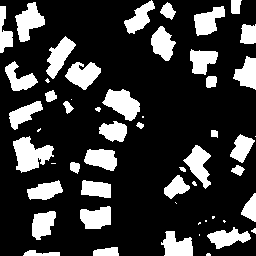} &
                \includegraphics[width=0.5\textwidth]{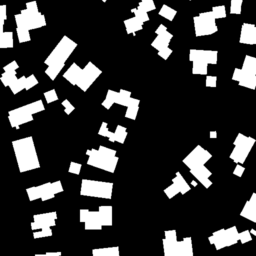} \\

                \includegraphics[width=0.5\textwidth]{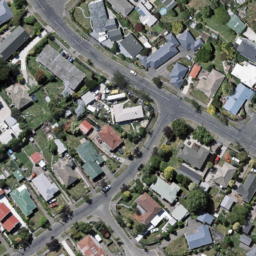} &
                \includegraphics[width=0.5\textwidth]{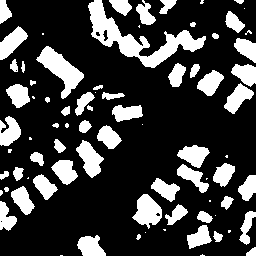} &
                \includegraphics[width=0.5\textwidth]{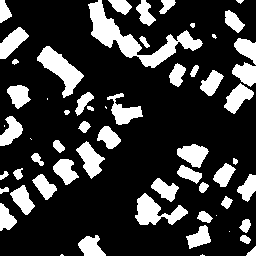} &
                \includegraphics[width=0.5\textwidth]{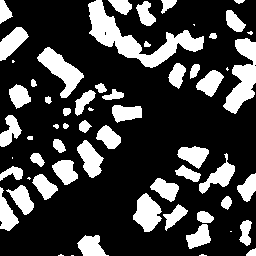} &
                \includegraphics[width=0.5\textwidth]{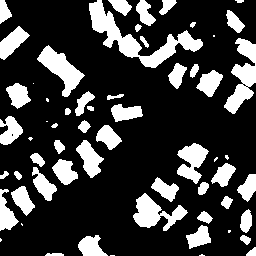} &
                \includegraphics[width=0.5\textwidth]{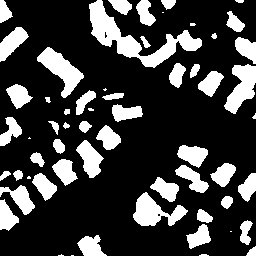} &
                \includegraphics[width=0.5\textwidth]{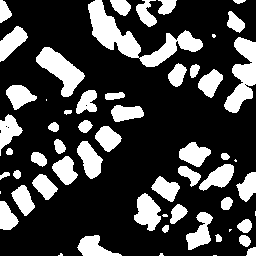} &
                \includegraphics[width=0.5\textwidth]{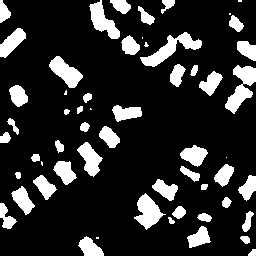} &
                \includegraphics[width=0.5\textwidth]{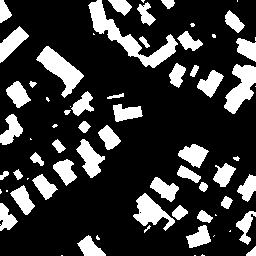} &
                \includegraphics[width=0.5\textwidth]{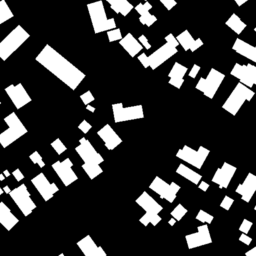} \\

            \end{tabular}
        }
        \caption{Semantic Segmentation: Qualitative comparison across all methods.}
        \label{fig:ss_progression_all}
    \end{subfigure}

    \vspace{0.5em}

    \begin{subfigure}{0.7\textwidth}
        \centering
        \resizebox{\textwidth}{!}{
            \begin{tabular}{cccccccccccc}
                \Huge Pre-Event & \Huge Post-Event & \Huge FCSiamConc & \Huge BIT & \Huge ChangeFormer & \Huge CGNet-CD & \Huge ELGC-Net & \Huge DDPM-CD & \Huge MambaBCD & \Huge Noise2Map (ours) & \Huge Ground Truth \\

                \includegraphics[width=0.5\textwidth]{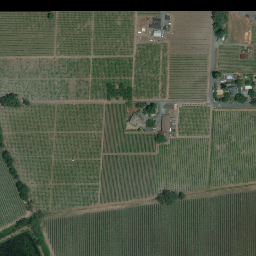} &
                \includegraphics[width=0.5\textwidth]{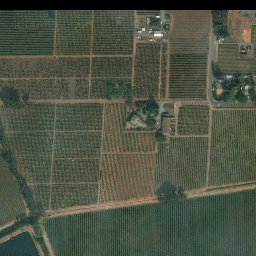} &
                \includegraphics[width=0.5\textwidth]{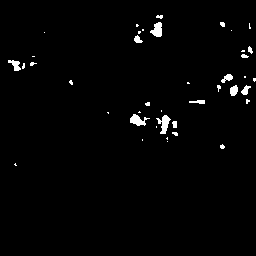} &
                \includegraphics[width=0.5\textwidth]{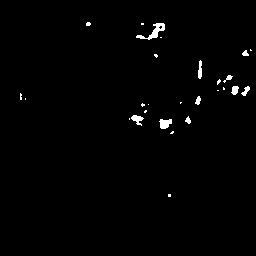} &
                \includegraphics[width=0.5\textwidth]{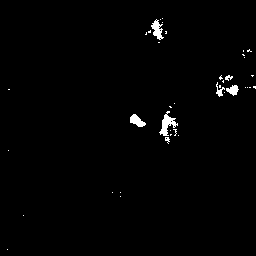} &
                \includegraphics[width=0.5\textwidth]{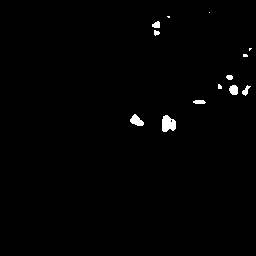} &
                \includegraphics[width=0.5\textwidth]{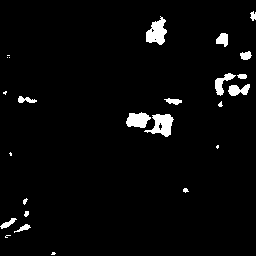} &
                \includegraphics[width=0.5\textwidth]{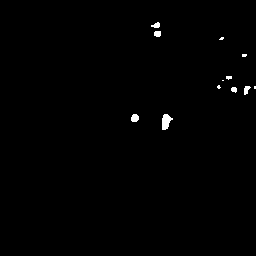} &
                \includegraphics[width=0.5\textwidth]{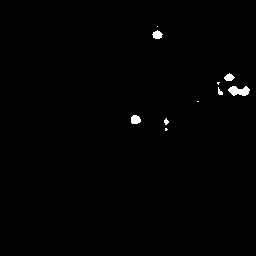} &
                \includegraphics[width=0.5\textwidth]{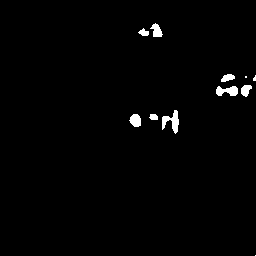} &
                \includegraphics[width=0.5\textwidth]{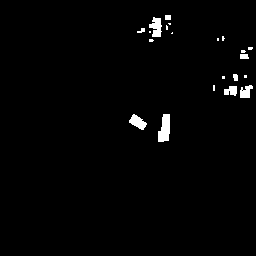} \\

                \includegraphics[width=0.5\textwidth]{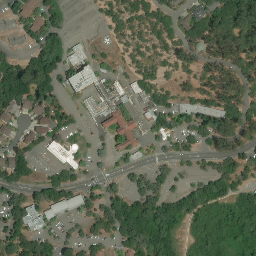} &
                \includegraphics[width=0.5\textwidth]{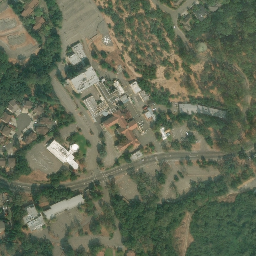} &
                \includegraphics[width=0.5\textwidth]{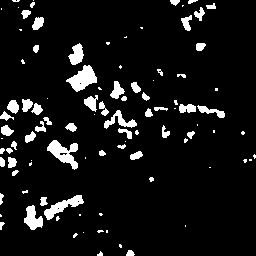} &
                \includegraphics[width=0.5\textwidth]{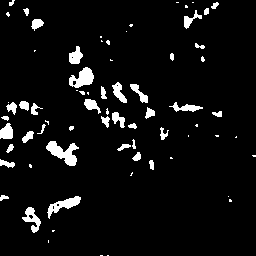} &
                \includegraphics[width=0.5\textwidth]{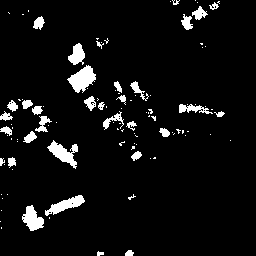} &
                \includegraphics[width=0.5\textwidth]{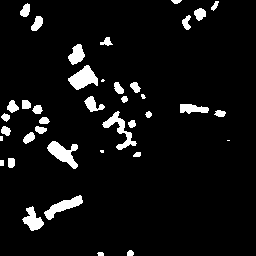} &
                \includegraphics[width=0.5\textwidth]{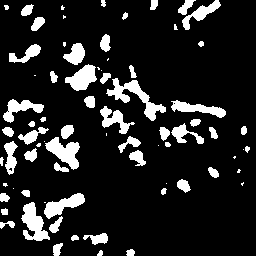} &
                \includegraphics[width=0.5\textwidth]{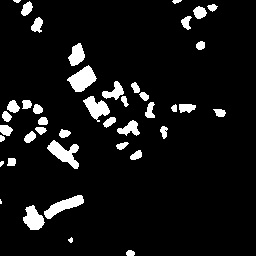} &
                \includegraphics[width=0.5\textwidth]{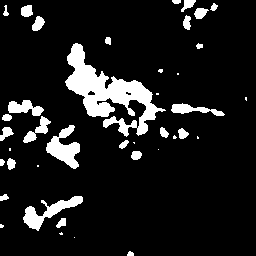} &
                \includegraphics[width=0.5\textwidth]{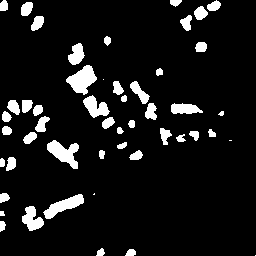} &
                \includegraphics[width=0.5\textwidth]{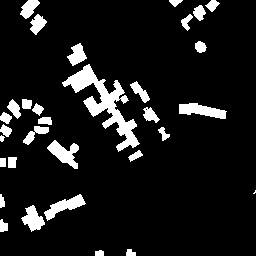} \\

            \end{tabular}
        }
        \caption{Change Detection: Qualitative comparison across all methods.}
        \label{fig:cd_progression_all}
    \end{subfigure}

    \caption{Qualitative comparison of change detection and semantic segmentation across all evaluated models. Our model \textit{Noise2Map} consistent demonstrates sharp boundaries and high accuracies as compared to the other models.}
    \label{fig:qualitative_comparison}
\end{figure*}

To complement the quantitative results, we perform qualitative analysis of Noise2Map on some input samples. Visual comparisons are provided for both CD and SS. As shown in Figure~\ref{fig:qualitative_comparison}, our model consistently generates sharp boundaries and accurate masks compared to other baselines. 
Noise2Map remains effective under moderate appearance variations because the noise scheduler exposes the model during training to multiple noisy representations supervised with the same change mask, encouraging focus on task-consistent changes rather than superficial appearance differences. For example, the Santa Rosa subset of xView2 includes off-nadir angles from approximately 5.7° to 22.8°, yet Noise2Map maintains strong performance, suggesting limited sensitivity to viewpoint variation.

\subsubsection{Interpretability Component}

\begin{figure}[t]
    \setcounter{subfigure}{0}
    \centering
    \setlength{\tabcolsep}{1.5pt}
    \renewcommand{\arraystretch}{0.92}

    \subfloat[SS: Diffusion process over timesteps.\label{fig:ss_progression}]{
        \begin{adjustbox}{max width=0.7\columnwidth} 
        \begin{tabular}{*{7}{C{0.14\linewidth}}}
            \scriptsize Original & \scriptsize $t{=}0$ & \scriptsize $t{=}10$ & \scriptsize $t{=}100$ & \scriptsize $t{=}500$ & \scriptsize $t{=}999$ & \scriptsize GT \\
            \img{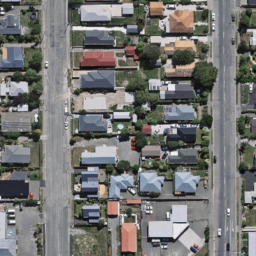} &
            \img{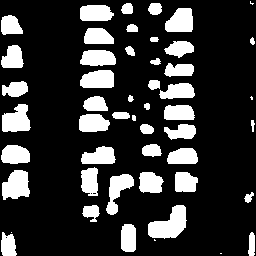} &
            \img{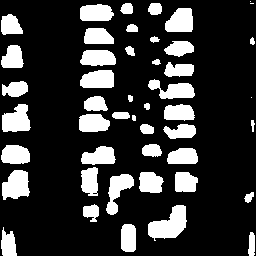} &
            \img{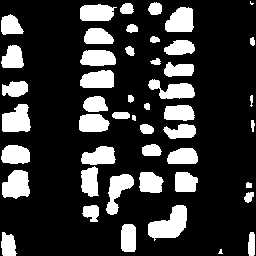} &
            \img{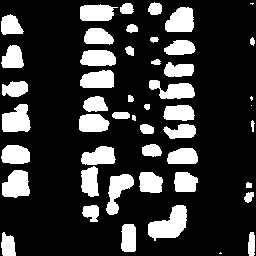} &
            \img{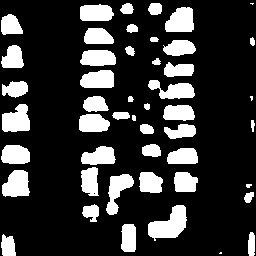} &
            \img{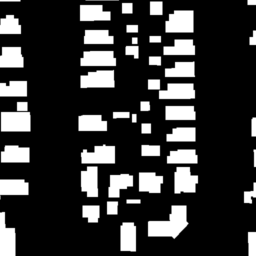} \\

            \img{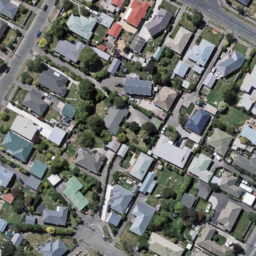} &
            \img{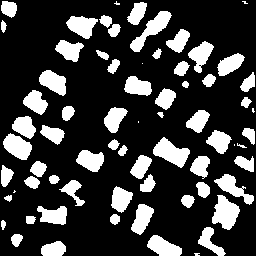} &
            \img{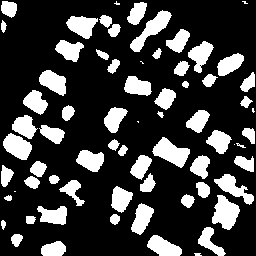} &
            \img{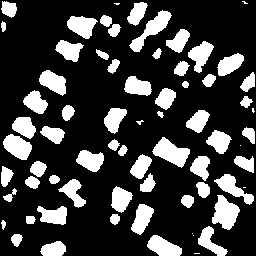} &
            \img{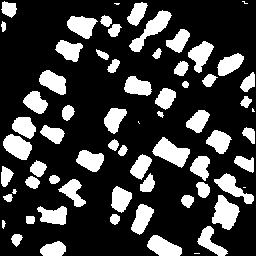} &
            \img{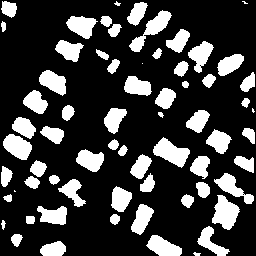} &
            \img{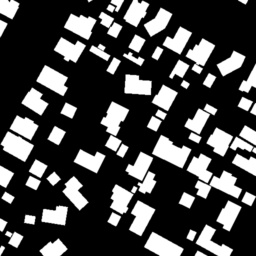} \\

            \img{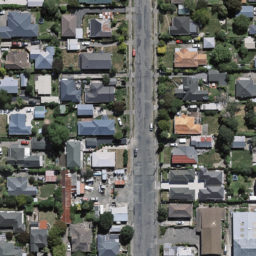} &
            \img{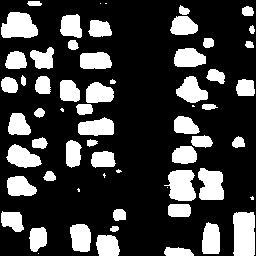} &
            \img{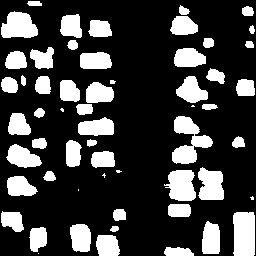} &
            \img{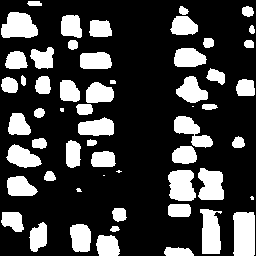} &
            \img{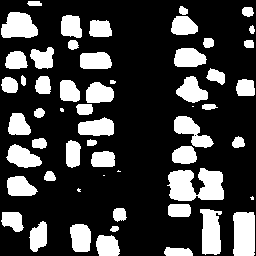} &
            \img{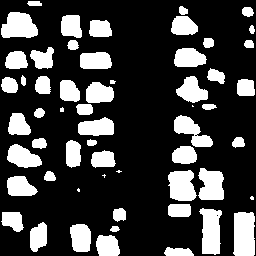} &
            \img{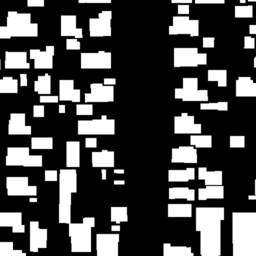} \\

            \img{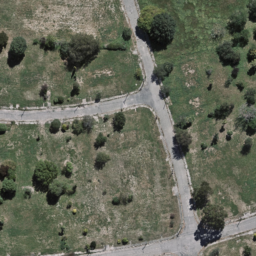} &
            \img{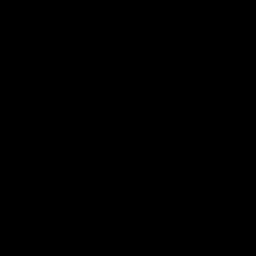} &
            \img{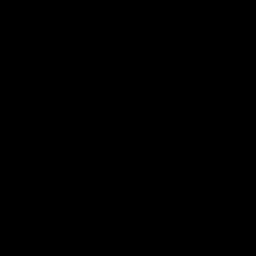} &
            \img{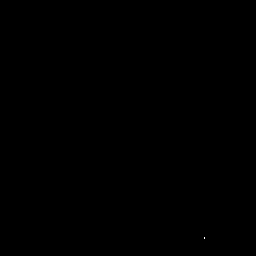} &
            \img{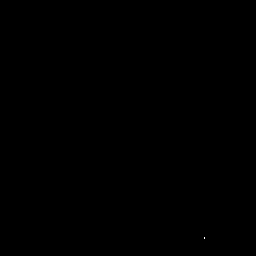} &
            \img{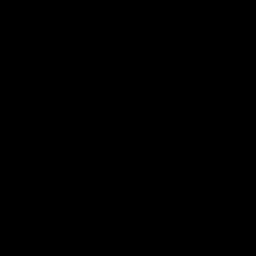} &
            \img{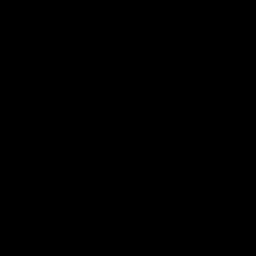} \\
        \end{tabular}
        \end{adjustbox}
    }


    \subfloat[CD: Diffusion process over timesteps.\label{fig:cd_progression}]{
        \begin{adjustbox}{max width=0.7\columnwidth} 
        \begin{tabular}{*{7}{C{0.14\linewidth}}}
            \scriptsize Pre-Event & \scriptsize Post-Event & \scriptsize $t{=}0$ & \scriptsize $t{=}10$ & \scriptsize $t{=}100$ & \scriptsize $t{=}500$ & \scriptsize GT \\
            \img{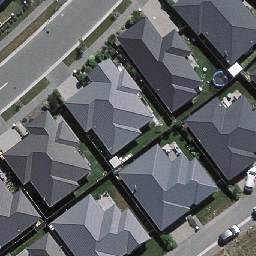} &
            \img{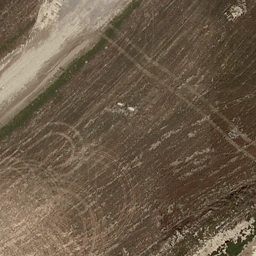} &
            \img{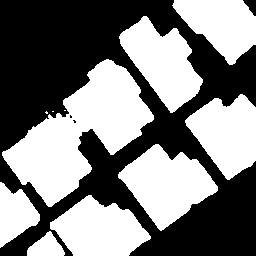} &
            \img{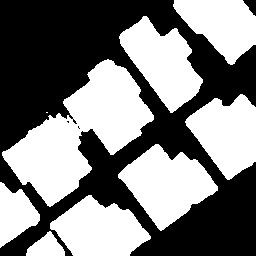} &
            \img{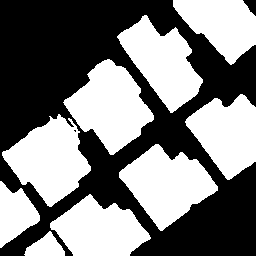} &
            \img{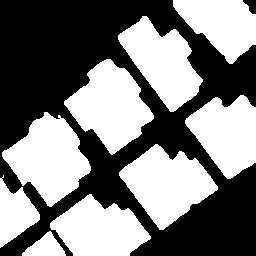} &
            \img{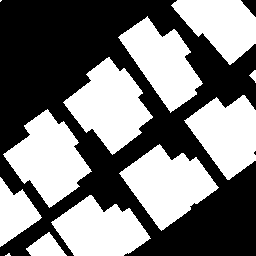} \\

            \img{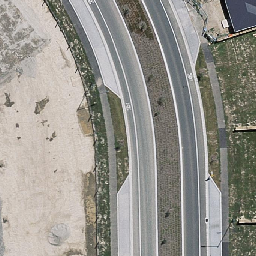} &
            \img{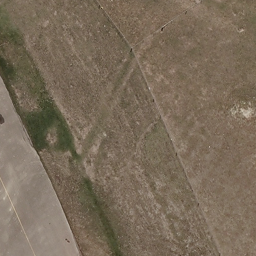} &
            \img{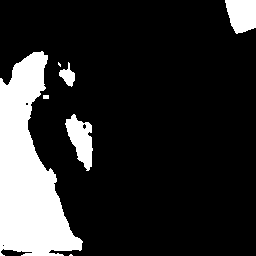} &
            \img{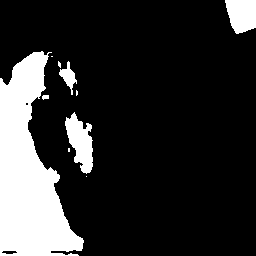} &
            \img{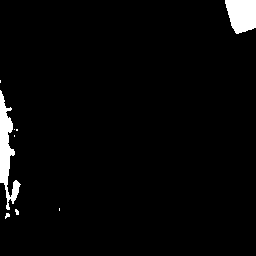} &
            \img{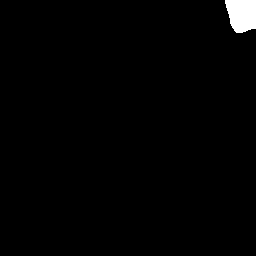} &
            \img{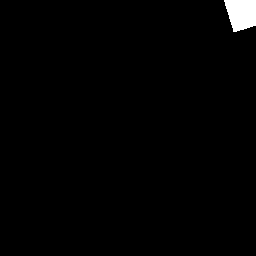} \\

            \img{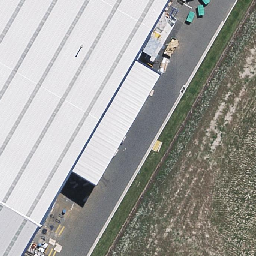} &
            \img{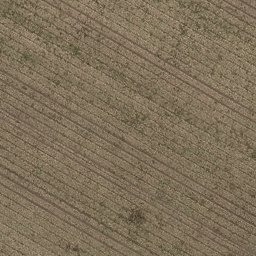} &
            \img{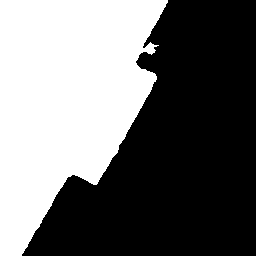} &
            \img{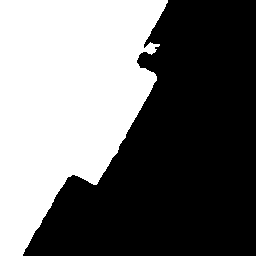} &
            \img{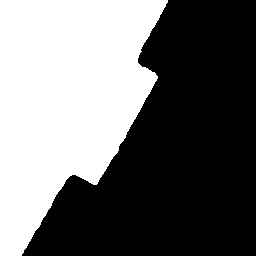} &
            \img{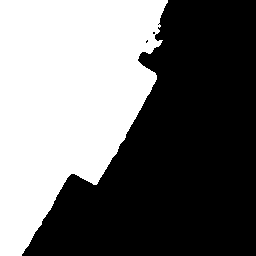} &
            \img{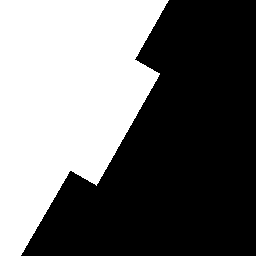} \\

            \img{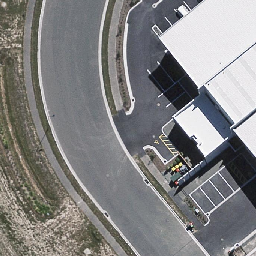} &
            \img{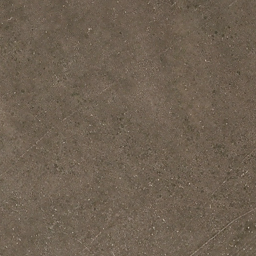} &
            \img{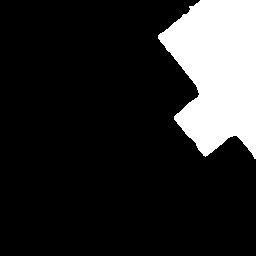} &
            \img{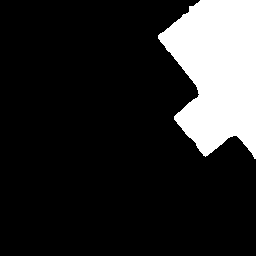} &
            \img{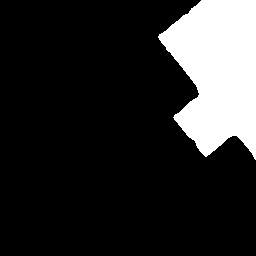} &
            \img{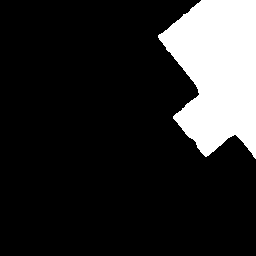} &
            \img{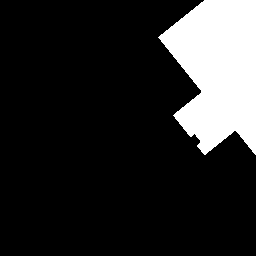} \\
        \end{tabular}
        \end{adjustbox}
    }

    \caption{Progression of predicted masks with Noise2Map over timesteps for SS (top) and CD (bottom).}
    \label{fig:timesteps_progression}
\end{figure}

Unlike conventional discriminative models that produce a single output without exposing intermediate reasoning steps, Noise2Map enables inspection of predictions across diffusion timesteps. Each timestep corresponds to a distinct diffusion-conditioned representation, allowing us to observe how semantic or change-related structures progressively emerge.
Figure~\ref{fig:timesteps_progression} visualizes the predicted segmentation and change maps at multiple timesteps, from early noisy states ($t=999$) to the final prediction ($t=0$). At high timesteps, predictions are coarse and fragmented, capturing only the most salient and spatially consistent regions. As the timestep decreases, finer structural details gradually appear, boundaries sharpen, and isolated false positives are suppressed. This behavior suggests that early diffusion stages encode global, low-frequency signals, while later stages refine high-frequency spatial details and object boundaries.

\begin{figure}[ht]
    \centering
    \includegraphics[width=0.6\linewidth]{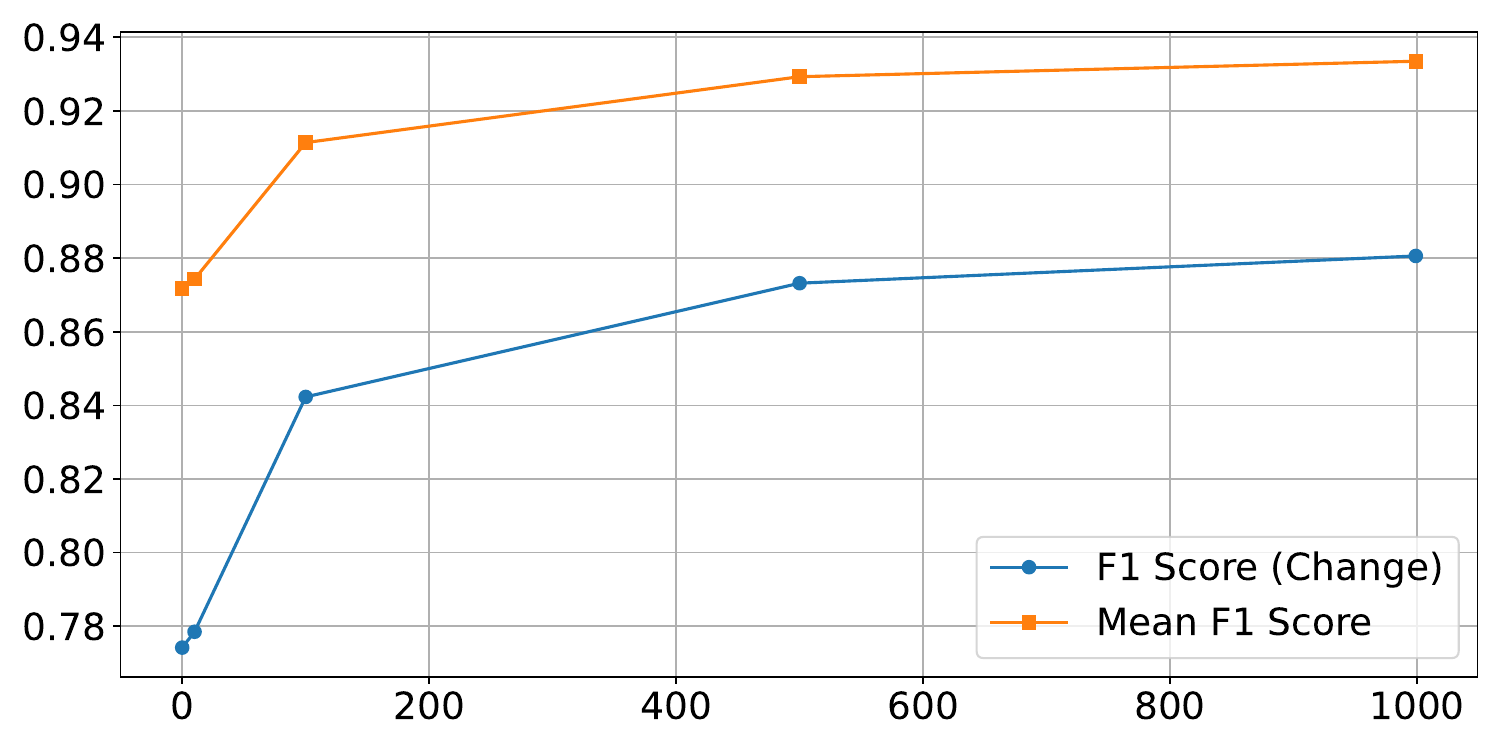}
    \caption{Evolution of F1 score during the reverse diffusion process. The F1 score for the change class (class 1) increases as the timestep decreases, showing that denoising progressively refines the model's predictions.}
    \label{fig:f1_vs_timestep}
\end{figure}

To quantify this behavior, we further compute the F1 score at various timesteps during the reverse diffusion process. As shown in Figure~\ref{fig:f1_vs_timestep}, the F1 score for the change class steadily improves over time until convergence, providing quantitative evidence of the model's progressive prediction refinement. We observe a clear increase in F1 score for the change class from 0.77 to 0.88 as the noise timestep decreases from $t = 999$ to $t = 0$, supporting our hypothesis that \textit{diffusion denoising progressively enhances discriminative signal quality}.

\subsection{Ablation Studies}

We investigate the impact of various design choices on the performance of Noise2Map. Specifically, we conduct ablation studies on the necessity of the discriminative diffusion formulation, the choice of pretraining dataset, the number of noising timesteps, the different noise schedulers, and multi-task learning with Noise2Map. 
For these experiments, we follow the same setup defined in \ref{training_setup}, except we train each model for 100 epochs while tracking validation loss to save the best checkpoints. We test the models on one of the datasets, WHU, which includes both change detection and semantic segmentation annotations, using 50\% of the data for training.

\subsubsection{Effect of Discriminative Diffusion Formulation}
To isolate the performance gains contributed by our proposed diffusion formulation, namely the structured noise process and timestep embeddings, we conducted a ``no diffusion'' baseline experiment. For this ablation, we utilized the exact same Attention-UNet backbone employed by Noise2Map but removed the timestep embeddings and the forward noising process. This effectively reduces the architecture to a vanilla encoder-decoder network trained directly to map clean inputs to the target masks. As shown in Table~\ref{tab:no_diffusion_ablation}, discarding the diffusion formulation leads to a drop in performance across both tasks. The F1 score drops by 3.06\% for semantic segmentation and 4.21\% for change detection. These results demonstrate that the strong performance of Noise2Map gains are driven by the diffusion noise process and timestep embeddings, which allows the model to leverage intermediate noisy states to capture multi-scale representations that a vanilla encoder-decoder struggles to learn.

\begin{table}[ht]
    \centering
    \caption{Effect of the Diffusion Component}
    \label{tab:no_diffusion_ablation}
    \resizebox{0.9\columnwidth}{!}{%
    \begin{tabular}{lcc}
        \toprule
        \textbf{Model Configuration} & \textbf{Semantic Segmentation} & \textbf{Change Detection} \\
        \midrule
        Noise2Map (W/o diffusion) & 89.06 & 78.72 \\
        \textbf{Noise2Map (W/ diffusion)} & \textbf{92.12} & \textbf{82.92} \\
        \bottomrule
    \end{tabular}%
    }
\end{table}

\subsubsection{Effect of Pretraining Dataset}\label{effect_of_pretraining_subsection}
We analyze the impact of the pretraining dataset on the performance of Noise2Label by comparing four initialization strategies: training from scratch and pretraining on ImageNet, MAJOR-TOM (Sentinel-2) \cite{francis2024major}, and AID. For a fair comparison, we sample $10{,}000$ images from each pretraining dataset. All models are subsequently fine-tuned under identical settings and evaluated using the F1 score, and results are shown in Table~\ref{tab:pretraining_effect}. On SS, training from scratch yields an F1 score of 78.01, while ImageNet pretraining improves performance to 82.11, demonstrating the benefit of generic visual representations. Pretraining on remote-sensing–specific datasets leads to further gains: MAJOR-TOM (Sentinel-2) achieves an F1 score of 89.09, and pretraining on AID results in the best performance with an F1 score of 92.12. On CD, ImageNet pretraining slightly degrades performance compared to training from scratch (73.62 vs. 76.33), whereas remote sensing–specific pretraining improves results, with MAJOR-TOM reaching 76.52 and AID achieving the best performance at 82.92. These results highlight the importance of domain-aligned pretraining for enhancing Noise2Label’s discriminative capability.

\begin{table}[ht]
    \centering
    \caption{Effect of different pretraining datasets} 
    \label{tab:pretraining_effect}
    \resizebox{0.9\columnwidth}{!}{%
    \begin{tabular}{lcc}
        \toprule
        \textbf{Pretraining Dataset} & \textbf{Semantic Segmentation} & \textbf{Change Detection} \\
        \midrule
        None (Scratch)              & 78.01 & 76.33 \\
        ImageNet (10k)              & 82.11 & 73.62 \\
        MAJOR-TOM (Sentinel-2, 10k) & 89.09 & 76.52 \\
        AID (10k) & \textbf{92.12} & \textbf{82.92} \\
        \bottomrule
    \end{tabular}%
    }
\end{table}

\subsubsection{Effect of Noising Timesteps}
The number of timesteps plays a crucial role in the model performance, as it determines the level of noise introduced and how effectively the model learns to reconstruct the data. Table~\ref{tab:ablation_timesteps} shows the impact of different timesteps on the performance of Noise2Map on each task, suggesting that an optimal range for this parameter exists depending on task. The results indicate that different tasks benefit from different numbers of timesteps. For CD, the highest performance is achieved with 1000 timesteps (82.92), with performance gradually decreasing as the number of timesteps increases. This suggests that fewer timesteps may provide a better balance between detail capture and noise removal in CD tasks, as too many timesteps could introduce excessive complexity, making it harder for the model to reconstruct meaningful information. In contrast, SS achieves the highest performance at 750 timesteps (90.74). This suggests that segmentation tasks may benefit from an intermediate number of timesteps that allows the model to capture sufficient detail without oversmoothing. Performance decreases slightly at 1000 timesteps, but interestingly, there is an improvement again at 1250 timesteps, though not surpassing the 750 timestep performance. These findings indicate that tuning timestep parameter is important per task.

\begin{table}[ht]
    \centering
    \caption{Effect of Noising Timesteps}
    \label{tab:ablation_timesteps}
    \resizebox{0.8\columnwidth}{!}{%
    \begin{tabular}{lcc}
        \toprule
        \textbf{Timesteps} & \textbf{Semantic Segmentation} & \textbf{Change Detection} \\
        \midrule
        500  & 88.74 & 82.52 \\
        750  & \textbf{90.74} & 81.03 \\
        1000 & 89.37 & \textbf{82.92} \\
        1250 & 89.99 & 80.39 \\
        \bottomrule
    \end{tabular}%
    }
\end{table}

\subsubsection{Effect of Noise Scheduler}
Furthermore, we check the robustness of our model when subject to different noise schedulers. The choice of noise scheduler is crucial, as it affects the quality of learned representations and predicted masks. We experiment with DDIM \cite{song2020denoising}, DDPM \cite{ho2020denoising}, PNDM \cite{karras2022elucidating}, and Heun \cite{karras2022elucidating} schedulers. Each scheduler has different characteristics: DDIM is efficient and requires fewer steps, DDPM provides a robust baseline but often requires more steps, PNDM improves computational efficiency and training stability with fewer steps, and Heun offers high stability. We use 1000 timesteps for each scheduler. Table~\ref{tab:ablation_schedulers} summarizes the results. The results show that our model performs consistently well across DDPM, DDIM, and PNDM schedulers, with a notable drop using Heun. This is likely due to Heun’s incompatibility with discrete-time training, as its continuous, second-order nature can introduce instability and misaligned denoising. In contrast, DDPM, DDIM, and PNDM are all designed for discrete-step diffusion and align well with the model’s training process. DDPM provides stable performance, DDIM offers faster inference with minimal loss in quality, and PNDM adds efficiency and smooth denoising. Among them, PNDM achieves the best SS score (90.55), while DDPM performs best for CD (81.96). This highlights the model’s robustness across compatible schedulers.

\begin{table}[ht]
    \centering
    \caption{Effect of Noise Schedulers}
    \label{tab:ablation_schedulers}
    \resizebox{0.8\columnwidth}{!}{%
    \begin{tabular}{lcc}
        \toprule
        \textbf{Scheduler} & \textbf{Semantic Segmentation} & \textbf{Change Detection} \\
        \midrule
        DDIM & 89.13 & 80.44 \\
        DDPM & 89.37 & \textbf{82.92} \\
        PNDM & \textbf{90.55} & 81.08 \\
        Heun & 61.60 & 55.00 \\
        \bottomrule
    \end{tabular}%
    }
\end{table}




\subsubsection{Multi-Task vs. Single-Task Learning}
We evaluate multi-task (MT) learning in Noise2Map by sharing one denoising U-Net backbone across CD and SS heads. The MT objective is a weighted combination of task losses:
\begin{equation}
\mathcal{L}_{\text{MT}} = \lambda_{\text{CD}}\mathcal{L}_{\text{CD}} + \lambda_{\text{SS}}\mathcal{L}_{\text{SS}},
\end{equation}
where $\mathcal{L}_{\text{CD}}$ and $\mathcal{L}_{\text{SS}}$ are weighted cross-entropy losses.

Using an unweighted sum ($\lambda_{\text{CD}}=\lambda_{\text{SS}}=1$), MT improves CD F1 (82.92 $\rightarrow$ 87.21) but degrades SS F1 (89.37 $\rightarrow$ 85.52), indicating negative transfer due to loss imbalance. Following this insight, we rebalance the MT loss weights. Starting with lower weights $\lambda_{\text{CD}}=\lambda_{\text{SS}}=0.5$, CD F1 improves (82.92 $\rightarrow$ 86.43) and SS F1 is restored (89.37 $\rightarrow$ 89.61). With $\lambda_{\text{CD}}=0.7$, $\lambda_{\text{SS}}=1.3$, CD F1 remains strong (82.92 $\rightarrow$ 86.19) while SS F1 further improves (89.37 $\rightarrow$ 90.65). These results show that the SS drop is not inherent to sharing the backbone; proper loss weighting restores and even improves SS performance while maintaining strong CD performance.

\section{Conclusion and Future Work}
\label{conclusion}

We introduced Noise2Map, an end-to-end diffusion model that repurposes the denoising trajectory for discriminative mapping in remote sensing. Unlike sampling-based diffusion methods, Noise2Map predicts semantic and change maps in a single forward pass using task-aligned noise schedules. It achieves top-ranked performance across both semantic segmentation and change detection among seven strong baselines per task. The diffusion process also provides interpretability, revealing how predictions progressively emerge across timesteps. We further showed that domain-aligned pretraining improves performance, highlighting diffusion models as effective discriminative learners beyond generative modeling. 

Future work will explore larger and multimodal pretraining datasets and extend the framework to additional tasks such as temporal progression modeling.

\section*{Acknowledgment}

This research is part of the EO-AI4GlobalChange project funded by Digital Futures, Stockholm, Sweden.

\ifCLASSOPTIONcaptionsoff
  \newpage
\fi



\bibliographystyle{IEEEtran}
\bibliography{bibtex/bib/main}
\end{document}